\documentclass[oneside,a4paper,english,links]{amca}
\usepackage{algorithm}
\usepackage{graphicx}
\usepackage{amsmath,amsfonts}

\newcommand{\set}[1]{\left\{#1\right\}}

\renewcommand{\vec}[1]{{\mathbf{#1}}}
\newcommand{\tensor}[1]{{\mathbf{#1}}}

\newcommand{\transpose}[1]{{#1}^{\rm T}}
\newcommand{\parent}{\mathrm{parent}}

\title{Anisotropic k-Nearest Neighbor Search Using Covariance Quadtree}

\author[ ]{Eraldo P. Marinho}
\author[ ]{Carmen M. Andreazza}
\affil[ ]{Universidade Estadual Paulista (S\~ao Paulo State University),
Departamento de Estat\'\i stica, Matem\'atica Aplicada e Computa\c c\~ao -- Rio
Claro, S\~ao Paulo, Brazil}

\begin{document}
\vspace{3cm}

\maketitle


\begin{abstract}
We present a variant of the hyper-quadtree that divides a multidimensional
space according to the hyperplanes associated to the principal components of
the data in each hyperquadrant. Each of the $2^\lambda$ hyper-quadrants is a data partition in
a $\lambda$-dimension subspace, whose intrinsic dimensionality $\lambda\leq d$ is
reduced from the root dimensionality $d$ by the principal components analysis,
which discards the irrelevant eigenvalues of the local covariance matrix.
In the present method a component is irrelevant if its length is smaller than,
or comparable to, the local inter-data spacing. Thus, the covariance
hyper-quadtree is fully adaptive to the local dimensionality.
The proposed data-structure is used to compute the anisotropic K nearest neighbors (kNN),
supported by the Mahalanobis metric. As an application, we used the present k nearest
neighbors method to perform density estimation over a noisy data distribution.
Such estimation method can be further incorporated to the smoothed particle hydrodynamics, allowing
computer simulations of anisotropic fluid flows.
\end{abstract}

\section{Introduction}\label{sec:1}

There is a number of problems in pattern recognition that requires efficient
algorithms to search for the k nearest neighbors (kNN) in multidimensional
feature spaces, \cite[e.g., ][]{mcnames01fast,379504,770370,1092803}.
For instance, color segmentation of images requires a kNN algorithm to select
 parts of the context
whose color range matches a query in the color space
\cite[e.g., ][]{DBLP:conf/accv/RehrmannP98}.
Algorithms for cluster analysis of multivariate data are better improved if
 based upon kNN techniques for
estimating both the class-independent and the class-conditioned probability
 densities \cite[e.g., ][]{2006TWB.51.513-525}. Moreover, many pattern recognition
problems might require density estimation techniques, which are in turn
efficiently performed if combining kNN algorithms with kernel interpolation,
which is a generalization of Parzen's windows \cite[e.g., ][]{Bishop1995,Duda}.

Pattern recognition techniques are applied even in contexts with concerns other
than of the usual information retrieval or scene analysis problems but as a
helper on solving the conservation equations in computer simulation of fluid
dynamics. For instance, {\em smoothed particle hydrodynamics, SPH}
\cite[e.g., ][]{1977MNRAS.181..375G,1977AJ.....82.1013L} is a kernel-based
technique to perform computer simulation of fluid dynamics using particles as
input vectors. The technique simulates the continuum by means of particle
interpolated quantities, using similar approaches of kernel based density
estimation.

Usually smoothed particle hydrodynamics codes require a kNN approach
to decide the kernel parameters
\cite[e.g., ][]{Marinho+Lepine2000,Marinho+Andreazza+Lepine2001},
and references therein, in order to perform both
density estimation and interpolated spatial derivatives of the fluid quantities
as, for instance, pressure gradient, stress-tensor divergence and Maxwell's
stress divergence. Also SPH is useful in plasma simulations as shown
in \cite{Jiang+Oliveira+Sousa2006}. Nowadays, SPH is an important particle
method used even in incompressible fluid simulations as presented in
\cite{Xu2009}. Recent works have focused the full adaptivity of the density
estimation technique of SPH concerning the anisotropy in interface regimes as
discussed formerly by \cite{Owen+Villumsen+Shapiro+Martel1998} and more
recently by \cite{Liu+Liu+Lam2006}.

We introduce the concept of covariance hyper-quadtree as an extension of the
traditional hyper-quadtree data structure, but now taking into account that the
orthogonal directions through which the space is subdivided into hyper-quadrants
are set by the covariance eigenvectors. Moreover, this novel approach
includes the principal components analysis method on the reduction of the data
dimensionality in each node.

Covariance trees are a relatively novel concept on hierarchical data
decomposition. However, we may find in the literature that such terminology is
somehow mismatching both on its purpose and in its data structure definition as
it has been shown for instance in the works of \cite{IROS04-BuHa,ma98modeling}.

From the \cite{IROS04-BuHa} view point, a covariance tree is a variant of the
k-D tree, in which the uniform orthogonal coordinate system is replaced by the
local coordinate system of the covariance eigenvectors, centered on the center
of mass of the data partition (node). In other contexts, covariance trees are
defined to map the strategies of multiscale stochastic processes \cite[e.g., ][]{Rib2006}.

A term nearly similar to covariance quadtrees was presented in the literature in
a previous work of \cite{Minasny2007}, but in a quite different approach, and
still preserving the traditional concept of quadtrees, in which the image
tessellation is taken along a fixed set of orthogonal directions.

In the present work, we get ride of the \cite{IROS04-BuHa} idea of covariance
tree, which is a strict binary tree, to introduce the concept of {\em covariance
quadtree} as a generalization of the traditional quadtree rectangular
image tessellation \cite{FinkelBentley1974}. In the present approach, each
data partition has its own local orthogonal coordinate system centered on the
local data expectation with the coordinate axes defined by the local covariance
eigenvectors, which in turn are the normal directions of the clipping
hyperplanes.

We apply the covariance quadtree to the kNN search using a bimodal metric, which
is a variant of both \cite{mcnames01fast} and \cite[hereafter
D'DR]{770370} scheme. In the D'DR method, the search data structure
is limited to a balanced binary tree, which maps the recursive split of each
data partition by the median hyperplane whose normal is the covariance's
principal component (see Figures 1 and 2 of D'DR). We extended
this interesting idea to the covariance quadtree, whose number of derived nodes
spans from 2 up to $2^D$ children, where $D$ is the space dimensionality.

The main purpose of the present paper is the presentation and validation of
an efficient algorithm of finding the exact K nearest neighbors from a query
vector in a multidimensional data space, either considering isotropic or
anisotropic searches. The concept of anisotropy is presently interpreted in
terms of the Mahalanobis metric. Since the k-nearest neighbors have an
ellipsoidal support, set by the covariance matrix for this region, the
anisotropic search requires a bootstrapping scheme of finding the optimal
k-nearest neighbors morphology. The applications of the present method for both
image processing and anisotropic SPH simulations shall be presented in future
works.

The paper is structured as follows. In Section~\ref{sec:2} is introduced
the concept of covariance quadtree, whose geometrical principles are discussed
in subsection~\ref{sec:2.1}, and the computational aspects of its data structure
are discussed in subsection~\ref{sec:2.2}. In subsection~\ref{sec:2.3} is
discussed the idea of intrinsic dimensionality which may change along the tree
construction if regarding each node as vector subspace spanned from the node
expectation, having the principal components as the orthonormal basis of the
subspace. The 2D image interpretation of the {covariance quadtree} is
illustrated with some examples in subsection~\ref{sec:2.4}, which helps the
reader to interpret correctly the main idea of the covariance quadtree. In
Section~\ref{sec:3} are discussed the methods of finding the isotropic
(subsection~\ref{sec:3.1}) and the anisotropic (subsection~\ref{sec:3.2}),
Mahalanobis based, k-nearest neighbors. The benchmark is discussed in
Section~\ref{sec:4} for the isotropic case. The anisotropic search has the same
time complex multiple of the isotropic case, which depends on the number of the
required iterations until convergence. One simple application for image
construction from noise images is shown in section~\ref{sec:application}.
Further considerations, perspectives and conclusions on the introduced method
are made in Section~\ref{sec:5}.

\section{Covariance quadtree} 
\label{sec:2} 

\subsection{Principles} 
\label{sec:2.1} 

\begin{figure}
  \center{
  \includegraphics[width=70mm]{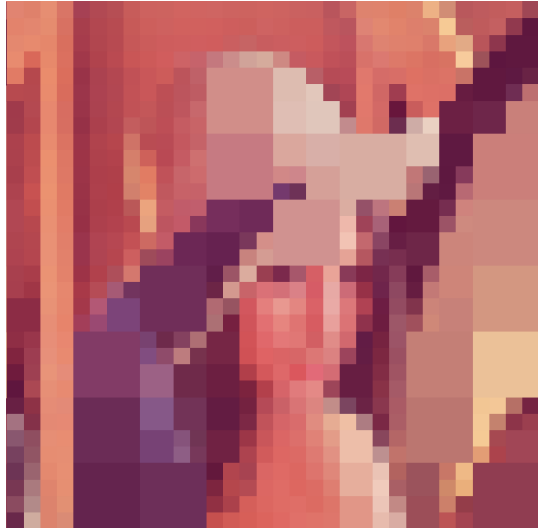}\\
  \caption %
	{The quadtree tessellated image of Figure~\ref{fig.t.1} for about
10\% tolerance in the gray-scale standard deviation.}
\label{fig.i.1} %
  } %
\end{figure}

\begin{figure}
\center{
\includegraphics[width=70mm]{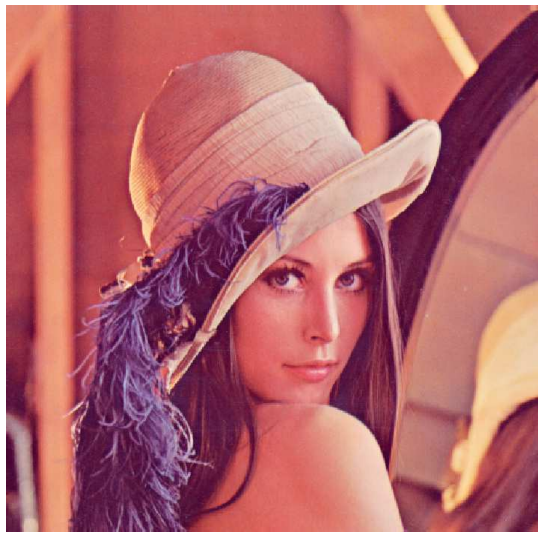}\\
\caption %
{The standard 512$\times$512, 24 bit, Lenna's portrait.}\label{fig.t.1} %
} %
\end{figure}

In order to introduce the concept of {covariance quadtree} we recall first to
the classical idea of a quadtree. Historically \cite{FinkelBentley1974} a
quadtree is the hierarchical data structure which maps the division of a
rectangle into four subrectangles, so that each internal node derives four child
nodes, and an external node derives a terminal partition, or image segment, which is
a rectangular atom of the image.
A simple example of a quadtree image tessellation is given in
Figure~\ref{fig.i.1} of the original image of Lenna shown in
Figure~\ref{fig.t.1}.

An immediate generalization of quadtree is its multidimensional form,
which divides a $d$-dimensional hyper-rectangle into $2^d$ child
hyper-rectangles, or hyper-quadrants. Hyper-quadtrees preserve the
aspect ratio of the the hyper-quadrants with the root.

For considerations of brevity, we assume hereafter that the term quadtree
applies both for the 2D case and for multidimensional case.

\begin{figure}
  \center{
  \includegraphics[width=70mm]{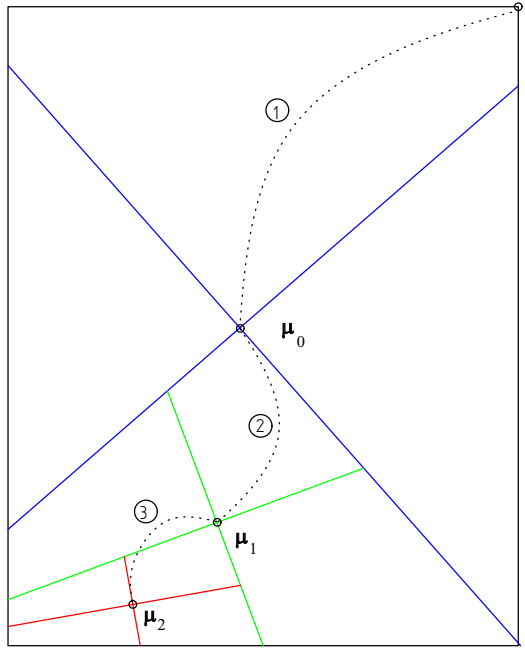}\\
  \caption %
	  {The first three steps of the top-down construction of a covariance  %
	  quadtree. The outer rectangle is the image frame, and the orthogonal %
	  strait lines, crossing the local expectation, are the principal      %
	  directions of the image segment. The dotted line illustrates the     %
	  top-down path.}\label{fig1}
  } %
\end{figure}

A {covariance hyper-quadtree}, or simply covariance quadtree, is a generalization of the traditional quadtree, with
the difference that covariance hyperplanes cuts the data space through the data
expectation, as sketched in Figure~\ref{fig1}. Henceforth, we call such a
spatial decomposition as {\em covariance tessellation} to make distinction from
the classic quadtree tessellation.

The {covariance quadtree} is the Gaussian multivariate extension of the
traditional quadtree data structure, but now with the difference that each node
spans from the local expectation $\vec\mu$ a vector space $\mathbb S$, whose
basis is the set of the local principal eigenvectors $\set{\vec v_1,\dots,\vec
v_{\lambda}}$. Any input vector $\vec x$ is then written down as $\vec
x=\vec\Delta+\vec\mu$, where $\vec\Delta$ is a vector in $\mathbb S$. Moreover,
each covariance node carries out the local aspect ratio in the form of the
$\lambda$ principal eigenvalues $\set{\alpha_1,\dots,\alpha_{\lambda}}$, which
by their own define the local data anisotropy as well as its intrinsic
dimensionality $\lambda$, where $\lambda$ is the number of principal components
of the covariance matrix.

Each of the principal covariance eigenvectors sets up an orthogonal hyperplane
passing through the expectation $\vec\mu$. Such a hyperplane is called a
covariance hyperplane, and for consistency the subspace $\mathbb S$ is the
covariance subspace. Each covariance hyperplane $\Pi_{\vec v}$ splits $\mathbb
S$ into two adjacent regions having $\Pi_{\vec v}$ as interface and the
expectation $\vec\mu$ as a common vertex. Thus, the $\lambda$-covariance
hyperplanes divide the covariance subspace into $2^\lambda$ partitions or
hyper-quadrants, which are assigned to $2^\lambda$ child nodes.

Each internal node obeying some division condition is subdivided by the
covariance hyperplanes into $2^\lambda$ children. The covariance hyperplane is
defined as the hyperplane which crosses the expectation and is directed by some
of the covariance eigenvectors. as its normal direction. The number $\lambda$ of
principal components is called the intrinsic dimensionality of the local data
distribution.

If the number $n$ of input vectors associated to the node is smaller than the
original data dimensionality $d$ then the intrinsic node dimensionality
$\lambda$ is limited to $0\leq\lambda\leq n-1$.

\subsection{Tree build} 
\label{sec:2.2} 

\subsubsection{Covariance quadtree data structure} 

The {covariance quadtree} is a hierarchical data structure formed by nodes whose
attributes are the following:
\begin{enumerate}
		\item a data subset represented as a list of the input vectors
		embedded
		in the space partition;\label{item1}
		\item the expectation and the $\lambda$-principal components
		evaluated
		over the local data content;\label{item2}
		\item a lower-bounding corner formed by the parent-evaluated
		expectation
		and the principal components renormalized in order to point
		outward
		the space partition;\label{item3}
		\item a link to visit the $2^\lambda$ possible
		children.\label{item4}
\end{enumerate}

The parental link is implemented by assigning a $2^\lambda$-dimensioned array
of pointers to child nodes. Given the data set $\mathbb{X}_\nu$, which
is assigned to the node $\nu$, there are $2^\lambda$ disjoint subsets
$\mathbb{X}_{\nu_\beta}$, each of which is assigned to the respective child
node $\nu_\beta$.

Each of the $n_\nu$ input vectors in $\mathbb{X}_\nu$ is assigned (or
moved) to one of the $2^\lambda$ partitions $\mathbb{X}_{\nu_\beta}$
$(\beta\in\mathbb{Z}_{_{2^\lambda}})$, which means that there is a
map $\beta:\mathbb{R}^\lambda\mapsto\mathbb{Z}_{_{2^\lambda}}$ to
derive a childhood from the interior node $\nu$:
\[
\mathbb{X}_\nu\underrightarrow{\;\;\;\;\beta\;\;\;\;}(
\begin{array}{ccc}
  \mathbb{X}_{_0} & \dots & \mathbb{X}_{_{2^\lambda-1}}
\end{array})
\]
where $\mathbb{Z}_{_{2^\lambda}}=\set{0,\dots,2^\lambda-1}$.

We adopt the following hash function, namely the hyperquadrant classifier
or child index, to map a input vector from its origin data set $\mathbb{X}_\nu$
to a (child) data partition $\mathbb{X}_{\nu_\beta}$:
\begin{equation}\label{eq:child-index}
\beta(\vec\Delta)=
\sum_{j=1}^\lambda\;2^{\lambda-j}\;H(\transpose{\vec\Delta}\;\vec
v_j),
\end{equation}
where $\vec\Delta$ is the data deviation about the expectation
$\vec\mu_\nu$, $\vec{v}_j$ is the $j$-principal eigenvector, and $H$
is the following step function:
\begin{equation}\label{eq:step-function}
H(x)=
\left\{
  \begin{array}{ll}
		1, & \hbox{$x\geq0;$} \\
		0, & \hbox{$x<0.$}
  \end{array}
\right.
\end{equation}

\begin{figure}
  \center{
  \includegraphics[width=70mm]{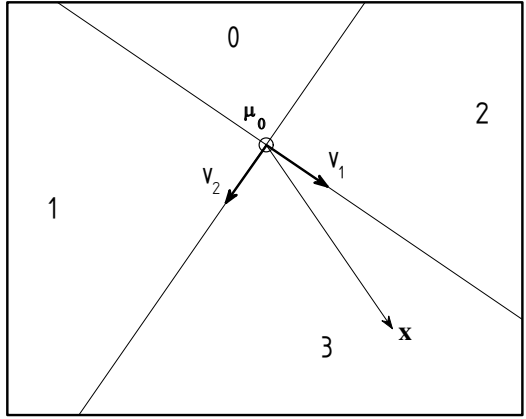}\\
  \caption %
		  {Child numbering order with respect to the principal
		  components detected in the dividing node according
		  to equations~(\ref{eq:child-index}) and
(\ref{eq:step-function}).
		  }\label{fig3} %
  } %
\end{figure}

A 2D instance of the child indexing scheme can be seen in Figure~\ref{fig3}.

After successive divisions each node is bounded not only by its lower-bounding
corner but also by the lower-bounding corners from all of its ancestors
but the root. The latter can be unbounded unless for the sake of the problem
details which can impose an input frontier as e.g. the frame illustrated
in Figure~\ref{fig1}. In this case, any interior node is wrapped by a convex
hull, formed by the innermost of the corners collected from all of the node
ancestors. As a result, the node boundary is then formed by the innermost
hyperplanes from both the local-lower bounding corner and the parent boundary.
The node boundary is then a synthesized attribute combined from the inherited
parent boundary and from the lower bounding corner.

In multidimension spaces it is somehow entangling to assemble a data structure
to represent the node boundary. However, in 2D-spaces it is too simple to
determine the boundary in terms of minimal polygons formed by the straight
lines orthogonal to the principal components.

If the root is unbounded, its childhood is lower-bounded by infinite
hyperpyramids having the root expectation as their common vertex. An
important result, not proven here, is that if the root is either unbounded
or it has a convex boundary, any interior node boundary is either an infinity
hyperpyramid or a convex hyperpolyhedron.

Since the set of covariance hyperplanes forms $2^\lambda$ lower bounding
corner with the expectation as the common vertex, each of these principal
hyperplanes isolates pair of children. Thus, each child has $\lambda$ common
face siblings. A 2D sketch of the covariance-quadtree tessellation was given
in Figure~\ref{fig1}. From the child point of view, any interior input vector
faces the concave part of the lower bounding corner, which requires a
direction multiplier $\phi\in\set{-1,+1}$ in order to have the child normal
vectors ever pointing outward its region.

From equation~(\ref{eq:child-index}), we have that the leftmost bit
of the child index $\beta$ (unsigned integer) represents the data
point orientation with respect to the principal hyperplane
($\vec{v}_{_1}$-direction): $0$, if the eigenvector is pointing
outward the data region, and $1$, if $\vec{v}_{_1}$ is pointing
inward. Thus, the direction multiplier $\phi_j$, as a function on
both the $j$-principal eigenvector and the child index $\beta$, is
given by
\begin{equation}\label{eq:direction-bit}
\phi_j=(-1)^{\;{\rm bit}(j,\;\beta)},
\end{equation}
where the function ${\rm bit}(j,\;\beta)$ returns the $j^{\;\mathrm{th}}$
left-to-right bit of the child index $\beta$.
Consequently, for each eigenvector $\vec{v}_j$, given the child
index $\beta$, corresponds the lower-bounding corner normal vectors
$\vec{u}_{\beta_j}$ given by
\begin{equation}\label{eq:normal-vector}
		\vec{u}_{j}=\phi_j\;\vec{v}_j
\end{equation}

\begin{figure}
\center{
\includegraphics[width=70mm]{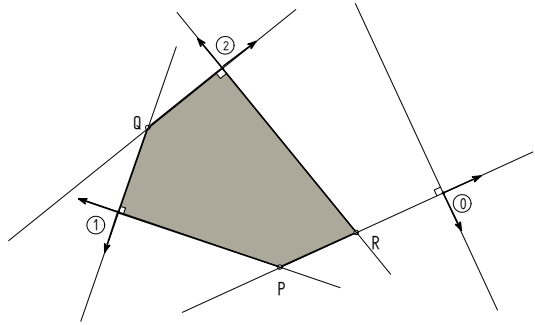}\\
\caption %
	{A simplified situation in which the node boundary is
assembled    %
	by combining the parent boundary with the local lower-
	bounding corner. The parent boundary is an inherited
attribute    %
	from its ancestors which is presented in the form of a
polygonal. %
	The lower-bounding corner intercepts some straight lines of
the   %
	parent boundary which has at least two solutions. The
solution    %
	is the innermost one which is faced by the concave part of
the    %
	lower-bounding corner.
} %
}\label{fig2}
\end{figure}

\subsection{Intrinsic dimensionality}\label{sec:2.3}

An interesting aspect of the {covariance quadtree} is that its spatial
tessellation keeps track of the residual covariance left on each child
partition. This is an interesting development since it reveals the natural
adaptivity of the decomposition method to the partition's intrinsic
dimensionality, yielded from the principal components analysis, PCA, aka
{\em the Karhunen-Lo\`eve transform} \cite[e.g., ][]{Jollife1986,Bishop1995}.

The proposed decision rule for the principal components analysis is that the
next training component, which is ever smaller than or equal to the
previous component, must be greater than the mean data spacing interior to the
hyper-rectangle having as edges the square root of the previously estimated
components.

The intrinsic dimensionality is decided either by limiting the number of
relevant components, or by spatial singularities such a data distribution having
no dispersion in some dimensions as the case of the local data being entirely
distributed in a surface. As the number $n$ of input vectors of a given
partition becomes smaller than the original data dimensionality $d$, the
intrinsic dimensionality $\lambda$ is geometrically constrained to $\lambda\leq
n-1$. Thus, a two-point partition has intrinsic dimensionality \textsc{one}
($\lambda=1$).

Top-level space partitions (nodes) likely have higher intrinsic
dimensionality than lower-level partitions. Moreover, our space division is
performed through the expectation, which implies that the division centers
(node vertices) tend to be closer to the denser child partitions than to the
rarefied ones.

\subsection{Two-dimensional case: image examples}\label{sec:2.4}

\begin{figure}
  \center{
  \includegraphics[width=70mm]{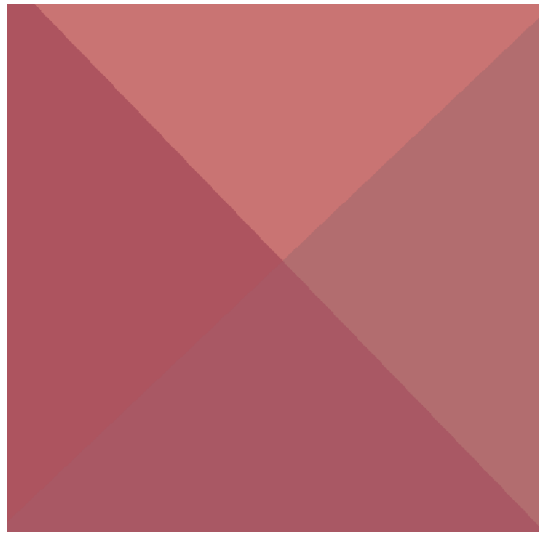}\\
  \caption %
		  {The first level covariance quadtree partition from the
original image in
		  Figure~\ref{fig.t.1}. Each partition is filled up with the
mean RGB color.
		  Compare this image with the sketch in Figure~\ref{fig1}.
		  }\label{fig.t.2} %
  } %
\end{figure}

\begin{figure}
  \center{
  \includegraphics[width=70mm]{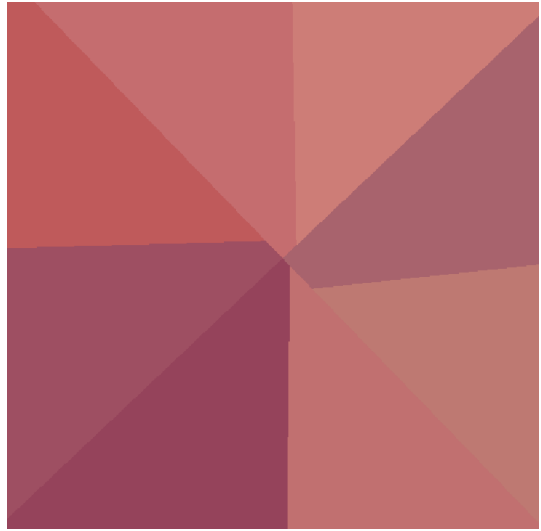}\\
  \caption %
	{
	The second level covariance quadtree partition from the original image
in
	Figure~\ref{fig.t.1}. The local dimensionality is reduced if the minor
component is below 50\% the
	major one.
	}\label{fig.t.3a} %
  } %
\end{figure}

\begin{figure}
  \center{
  \includegraphics[width=70mm]{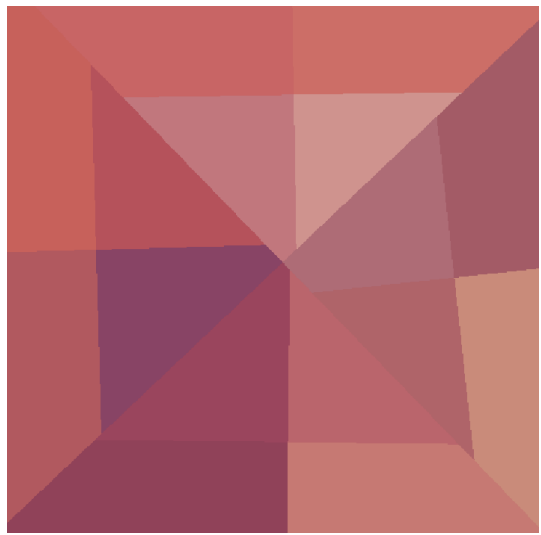}\\
  \caption %
		  {As in previous figure, but with no
dimensionality detection.
		  }\label{fig.t.3b} %
  } %
\end{figure}

\begin{figure}
  \center{
  \includegraphics[width=70mm]{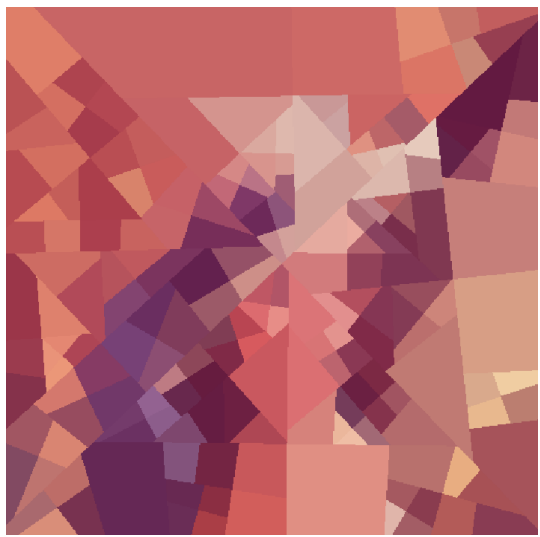}\\
  \caption %
		  {Level 4 covariance quadtree, within 50\% tolerance for the
dimensionality detection.
		  }\label{fig.t.4} %
  } %
\end{figure}

To better illustrate the {covariance quadtree} datastructure, we three examples
of
image tessellation by a hierarchy of
grey scale covariance. in this case, the root image is divided into 2 or 4
partitions, depending on the dimensionality reduction criterion applied.

In the following examples we assume a simple dimensionality reduction scheme,
where
an image partition is considered 2D if the minor-to-major component ratio is
above
a prefixed threshold, and then is divided into 4 partitions directed by the
eigenvectors.
Otherwise, the partition is cut perpendicularly to the principal component.

\begin{figure}
  \center{
  \includegraphics[width=70mm]{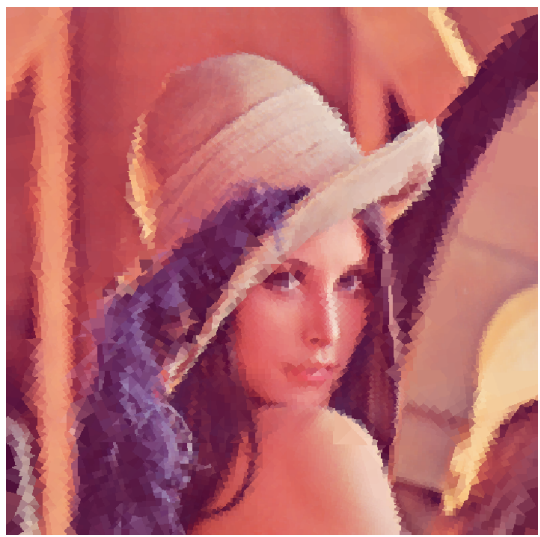}\\
  \caption{Resuming from the tree situation shown in Figure~\ref{fig.t.4},
until Level 7 in the covariance quadtree, also adopting 50\% tolerance for the
dimensionality detection.}\label{fig.t.5} %
  } %
\end{figure}

Each of the $\lambda$ hyperplanes from the lower-bounding corner intercepts
other $\lambda$ hyperplanes from the parent boundary to determine a vertex
on the $\lambda$-dimension hyperpolyhedron. In the particular case of 2D images,
as illustrated in Figure~\ref{fig.t.1}, we have the situations depicted in
Figures~\ref{fig.t.2} through \ref{fig.t.5}.

Figure~\ref{fig.t.2} illustrates the first level in the {covariance quadtree}
for Figure~\ref{fig.t.1}, assuming no dimensionality threshold used on choosing
the covariance principal components. Each image partition is filled up with the
mean RGB-color. The descent stop condition is that the partition's greyscale
dispersion is below a fraction (tolerance parameter) of the parent's greyscale
dispersion.

Figure~\ref{fig.t.3a} resumes the image tessellation from the previous
Figure~\ref{fig.t.2}, but assuming a simple dimensionality reduction. If the
minor-to-major components ratio is smaller than 0.5, the dimensionality is
reduced to {\sc one} ($d=1$).

Similar to Figure~\ref{fig.t.3a}, Figure~\ref{fig.t.3b} resumes the image
tessellation from the previous Figure~\ref{fig.t.2}, but assuming no
dimensionality reduction ($d=2$, regardless the tree depth).

Resuming from Figure~\ref{fig.t.3a}, we have the {covariance quadtree} situation
for level 4 shown in Figure~\ref{fig.t.4}. Now, some partitions are considered
1D, and other ones are considered 2D, depending on the aspect ratio of each
resulting partition. Finally, in the level 7 of the {covariance quadtree}, we
have the tessellated image shown in Figure~\ref{fig.t.5}.

\section{Covariance-based kNN}
\label{sec:3}

\subsection{Isotropic search}
\label{sec:3.1}

The present covariance-based K nearest neighbors algorithm is a top-down search that recursively proceeds 
the descent through nodes which may have some point closer than any of the k ($\leq K$)
collected nearest candidates until that level in the tree.

The adopted query-to-node distance is one of many forms of expressing distance
of a point to a set. To compose the query-to-node distance, it is first required
to define a pseudo distance, namely the direct query-to-node distance $D(.)$ as
the following function:
\begin{equation}\label{eq9}
D(\vec{x},\verb!node!)=\max_{j=1}^{\lambda}\;\bigl\{
				\transpose{(\vec{x}-\vec\mu)}\;\vec{u}_j
				\bigr\},
\end{equation}
where $\vec\mu$ is the node's lower-bound corner vertex and
$\set{\vec{u}_1,\dots,\vec{u}_\lambda}$ the set of its normal unit vectors as
discussed in Subsection~\ref{sec:2.2}. Since the internal product
$\transpose{(\vec{x}-\vec\mu)}\;\vec{u}_j$ is positive only if the query point
is facing the $j$-hyperplane from outside the node, the direct distance defines
a lower boundary for a collection of candidate vectors to comprise the list of
the k nearest neighbors.

Further examining equation~(\ref{eq9}) one may conclude that if the query vector
is facing the hyperplanes from inside, then $D(\vec{x},\verb!node!)\leq0$. The
equality occurs if, and only if, the query is lying in one of the node's
boundaries.

Since the covariance nodes but the root are bounded not only by its principal
hyperplanes but also by at least one of its ancestors boundary, we write the
complete query-to-node distance as the following recursive function:
\begin{equation}\label{eq8}
d(\vec{x},\verb!node!)=
      \max\left\{
		d(\vec{x},\parent(\verb!node!)),
		D(\vec{x},\verb!node!)
      \right\},
\end{equation}
with the breaking condition that $d(\vec{x},\verb!node!)=0$, if
$\verb!node!=\verb!root!$.

The distance calculation in equation~(\ref{eq8}) is performed along the tree
descent, and the returned value from the recursive call for $d(.)$ in the RHS is
an inherited attribute from the node's ancestor.

It sounds like a theorem that the distance from any query vector to the
covariance node is zero if the query is included in the node. In fact, the
direct distance $D(\vec{x},\verb!node!)$ is negative or null if the query is
interior or it is in some of the node's principal hyperplanes, which is valid to
the distance from the query to some of its ancestors until the root, where the
distance given by equation~(\ref{eq8}) is trivially zero. Since the maximum from
zero and a non-positive number is zero, the distance from the query to any node
which encloses it is zero.

\begin{algorithm}
\caption{The C-code for the recursive top-down kNN algorithm.}
\begin{verbatim}

1:  extern query_type q;
2:  extern list_type kNN_list;
3:
4:  void kNN_find (node_t *cell, double d)
5:  {
6:      d = query_to_node_distance (q, cell, d);
7:      if ( kNN_list.N == K && d > kNN_list.top_dist)
8:          return;
9:      else {
10:         if (cell->N > 1) {
11:             int  b;
12:             for (b = 0; b < cell->nchilds; b++)
13:                 if (cell->child[b])
14:                     kNN_find (cell->child[b], d);
15:         }
16:         else
17:             kNN_insert (cell, d);
18:     }
19: }

\end{verbatim}
\end{algorithm}

The present kNN scheme requires an upper-bounded neighbor list (kNN list), which
stores each input vector according to its query-to-data distance, so that the
end term is the outermost one from the query neighborhood. The closest neighbor
is of course the first element of the list after the complete search. Each input
vector is pointed to by a data descriptor, which is stored both in the neighbor
list as well as in the {covariance quadtree} nodes as previously discussed.

For the present, a neighbor is an input vector, which is assigned to a unit node
(leaf). The kNN list has \verb!K! members at most, where \verb!K! is the user
assigned number of nearest neighbors to find. The list is initially empty and it
progressively grows with the successive insertion of nearest neighbor candidates
along the descent until the list is fully populated with \verb!K! members.
Henceforth, any new insertion implies in removing the farthest element from the
list.

Positive query-to-node distance denotes the maximum of the distances from the
query to the hyperplanes bounding the node. On the other hand, the distance from
the query to a hyperplane is the Euclidean distance from the query to the
closest point in the hyperplane. Thus, if the farthest of the kNN candidates is
still closer than the focused node, then no closer neighbors can be found in
that node. This is the refutation criterion to be applied along the tree descent
if the kNN list is fully populated with the \verb!K! nearest candidates.

Conversely, a given node potentially has kNN candidates if either the kNN list
is not full yet, or the query-to-node distance is below or equal to the distance
from the query to the outermost candidate in the kNN list. Particularly in this
case, if the node is unit, the calculated distance is the Euclidean
point-to-point distance, and then the insertion method is called to solve in
what position in the list, if any, the new candidate might be inserted.

Both paragraphs above are summarized in the form a C code, listed in the
Algorithm~1. The proposed method is a recursive descent, with a prefixed call to
the query-to-node distance function given in equations~(\ref{eq9}) and
(\ref{eq8}). The second argument, \verb!d!, in \verb!kNN_find! is initially the
distance from the query to the parent node, which might be modified in the
assignment statement at line 6 conform equation~(\ref{eq9}).

According to the algorithm, the stop condition for tree descent is that the kNN
list is full and that the query-to-node distance is greater than the distant to
the outermost list member, ending the recursion through line 8. Line 10 decides
whether or not the algorithm recursively descends (lines 11-14) or it includes
the new neighbor candidate to the kNN list (line 17).

\begin{figure}
  \center{
  \includegraphics[width=70mm,height=70mm]{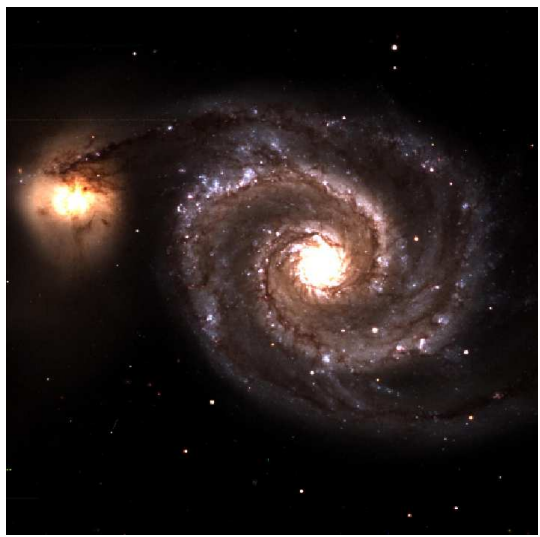}\\
  \caption %
{The original RGB image of the M51 galaxy, whose gray-scale (R+G+B) is
used to drive the random distribution shown in the next two
figures.}\label{fig.m51.full.1}
  } %
\end{figure}

\begin{figure}
  \center{
  \includegraphics[width=100mm,height=67.5mm]{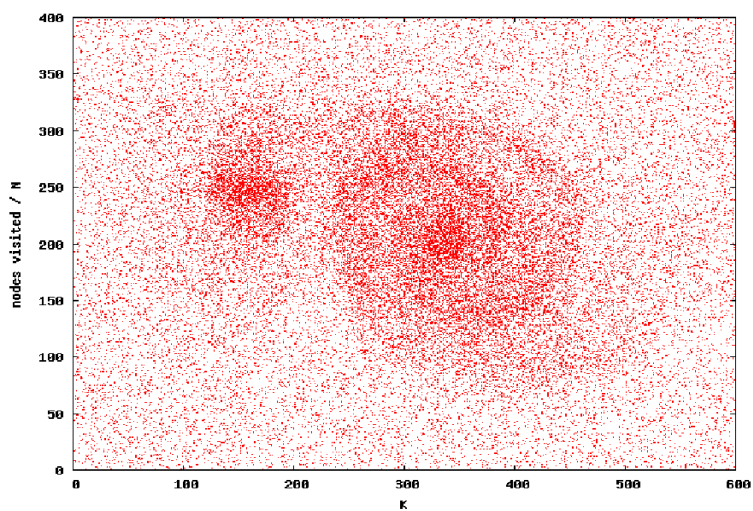}\\
  \caption %
	  {A Plot of 44,081 random points sampling the
gray-scale image of the galaxy M51.}\label{fig.m51.points.1}
  } %
\end{figure}

\begin{figure}
  \center{
  \includegraphics[width=100mm,height=67.5mm]{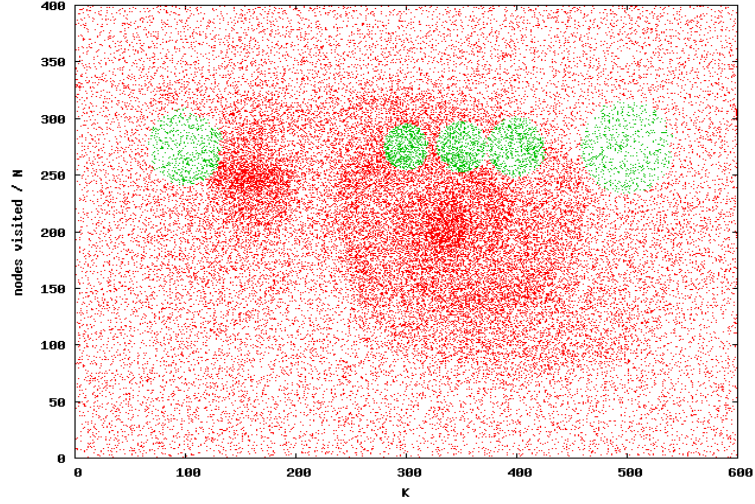}\\
  \caption %
	  {A plot of K=320 nearest neighbors (in green)  for 5
arbitrarily chosen points from previous figure.}\label{fig.m51.kNN.1}
  } %
\end{figure}

As an example of the search for exact kNN, we performed the search
on a random distribution of points which was drawn by the probability density
set proportional to the gray-scale distribution of
M51 spiral galaxy image (see Figure~\ref{fig.m51.full.1}), whose data points
are plotted in Figure~\ref{fig.m51.points.1}. Five query points were arbitrarily
chosen as shown in Figure~\ref{fig.m51.kNN.1}.

\subsection{Anisotropic search}\label{sec:3.2}

The algorithm discussed in previous section performs an isotropic
k-nearest neighbor search using the Euclidean metric. However, as the
covariance quadtree itself suggests, the search algorithm is worth
applied to detect a fixed number of the nearest neighbors accordingly to the
Mahalanobis metric \cite{Mahalanobis1936}, which is the natural choice to
find the nearest vectors accordingly to the covariance matrix estimated for the
neighborhood found.

The Mahalanobis distance $\xi$ from a query vector $\vec{x}$ to the
expectation $\vec\mu$ of a given data distribution explained by a covariance
$\tensor{\Sigma}$ is defined as the following positive definite bilinear form:
\begin{equation}
\xi^2=(\vec{x}-\vec\mu)^{\rm T}\tensor{\Sigma}^{-1}(\vec{x}-\vec\mu)
\label{eq:def:mahalanobis}
\end{equation}

In the present case, equation~(\ref{eq:def:mahalanobis}) is valid for a
data partition assigned to
as a {covariance quadtree} node. Thus, $\vec\mu$ is the node vertex about which
further
division might occur.

Let $\set{\vec{u}_j\,|\,j=1,\dots,D}$ be the set of all eigenvectors of the
covariance matrix, evaluated
for the K-nearest neighbors, and $\set{\sigma_j^2\,|\,j=1,\dots,D}$ its
eigenvalues. Then, the inverse of
the covariance matrix $\tensor{\Sigma}^{-1}$ may be rewritten in the diagonal
form:
\begin{equation}
\tensor{\Sigma}_{\rm kNN}^{-1}=\sum_{j=1}^D
\frac1{\sigma_j^2}\vec{u}_j\vec{u}_j^{\rm T}
\label{eq:diagonalcovariance}
\end{equation}

If regarding the principal components analysis to perform the
dimensionality reduction, the number of
relevant eigenvectors $\lambda$ might be smaller than the input space
dimensionality $D$, as discussed in subsection~\ref{sec:2.3}.

The query-to-point distance $\xi_{(\vec{x},\vec{x}_j)}^2$ is a modification
in the Mahalanobis distance so that
\begin{equation}
\xi_{(\vec{x},\vec{x}_j)}^2=(\vec{x}-\vec{x}_j)^{\rm T}\tensor{\Sigma}_{\rm
kNN}^{-1}(\vec{x}-\vec{x}_j)
\label{eq:query2point:mahalanobis}
\end{equation}

The distance from the query to a {covariance quadtree} node defined in
previous
subsection is just the maximum
Euclidean distance from the query to the node hyperplanes. The modification of
this distance to the Mahalanobis
metric is then
\begin{equation}
\xi_{(\vec{x},{\rm node})}^2= d_{(\vec{x},{\rm node})}^2
		\vec{u}_{j_{\rm max}}^{\rm T}\tensor{\Sigma}_{\rm
kNN}^{-1}\vec{u}_{j_{\rm
max}},
\label{eq:query-2-node:mahalanobis}
\end{equation}
where $d_{(\vec{x},{\rm node})}^2$ is the query-to-node distance calculated via
equation~(\ref{eq8}), and
$\vec{u}_{j_{\rm max}}$ is the node's normal vector that gave the maximum
contribution in evaluating
the RHS of the equation~(\ref{eq8}).

Algorithm~1 is robust regardless the metric chosen to classify the
covariance quadtree\ nodes as
kNN candidates. Thus, no changes are required in the algorithm of the previous
subsection. Only the distance
definition must be modified to the Mahalanobis metric to perform the anisotropic
search. Moreover, since we
do not know prior the covariance of the not yet known K nearest anisotropic
neighbors, the kNN modification requires bootstrapping.

The bootstrapping algorithm to perform the anisotropic search for the k
nearest neighbors under the
Mahalanobis metric works as follow:
\begin{enumerate}
\item Initialize the kNN covariance matrix with the identity matrix.
\item Repeat:
\begin{enumerate}
  \item Perform the kNN via Algorithm~1 with the distances modified according
      to equations~(\ref{eq:query2point:mahalanobis}) and
(\ref{eq:query-2-node:mahalanobis})\label{step:kNN};
  \item For the present kNN list, calculate the covariance matrix;
\end{enumerate}
\item Until convergence occurs on the covariance eigenvalues.
\end{enumerate}

Convergence occurs in few steps even if we adopt a tight tolerance as for
instance $10^{-19}$ used in the runs of present work.

\begin{figure}
  \center{
  \includegraphics[width=100mm,height=67.5mm]{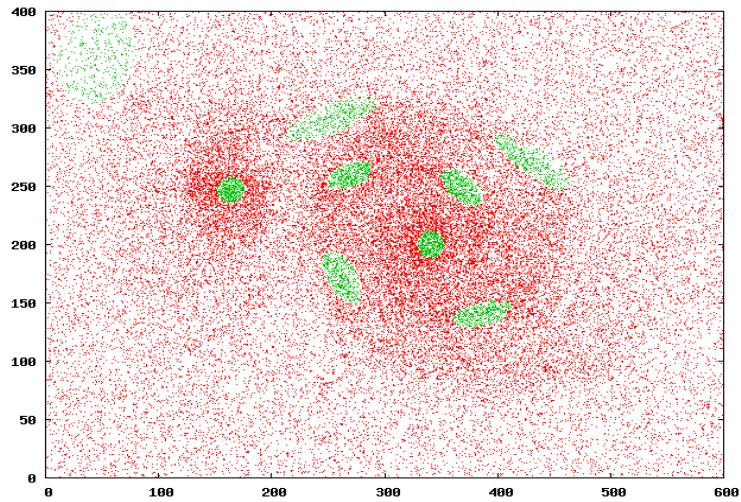}\\
  \caption %
	  {A plot of the K=410 nearest neighbors (in green) for 8
	  arbitrary queries for the data in figure~\ref{fig.m51.points.1}
	  but performing a Mahalanobis based anisotropic
	  search. Note that the green spots align with the preferential
	  directions of the data distribution.}\label{fig.m51.kNN.2}
  } %
\end{figure}

To illustrate the anisotropic search, we performed the method above for the
same data used in previous
subsection. Figure~\ref{fig.m51.kNN.2} shows the points plotted in green
corresponding to the $K=410$ neighbors found for 9 queries conveniently chosen
to reveal morphological aspects of the galaxy noise image. The leftmost
upper green spot is almost isotropic since corresponds to the image background.
However, in denser regions, the
kNN profiles follow the spiral pattern of the galaxy M51.

\section{Benchmark}
\label{sec:4}

\begin{figure}
  \center{
  \includegraphics[width=100mm,height=100mm]{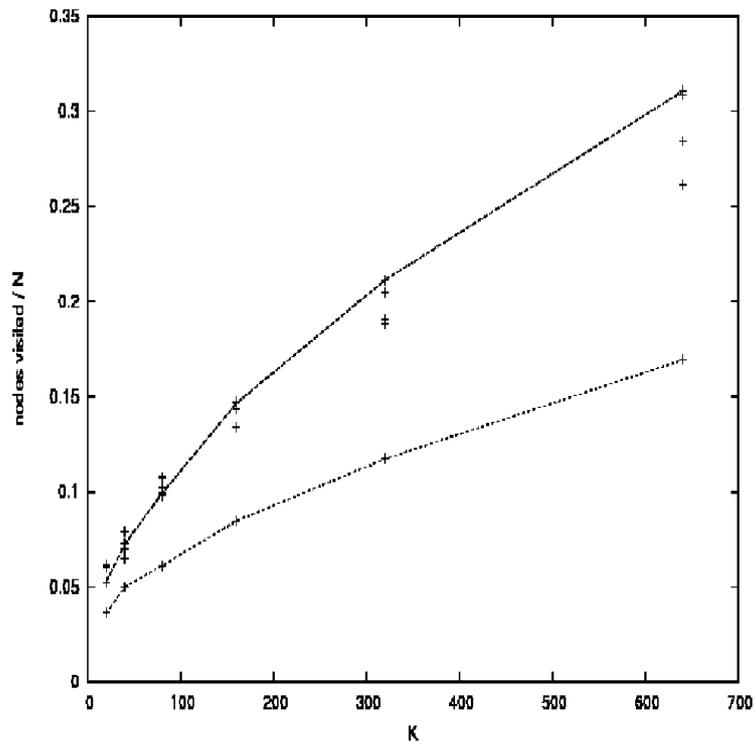}\\
  \caption %
  {Number of nodes visited per data point versus the number of K nearest
  neighbors found. Upper and lower curves are plotted for the maximum and
  minimum number of candidate nodes found, respectively.}\label{fig.timing.1}
      } %
\end{figure}

\begin{figure}
  \center{
  \includegraphics[width=100mm,height=100mm]{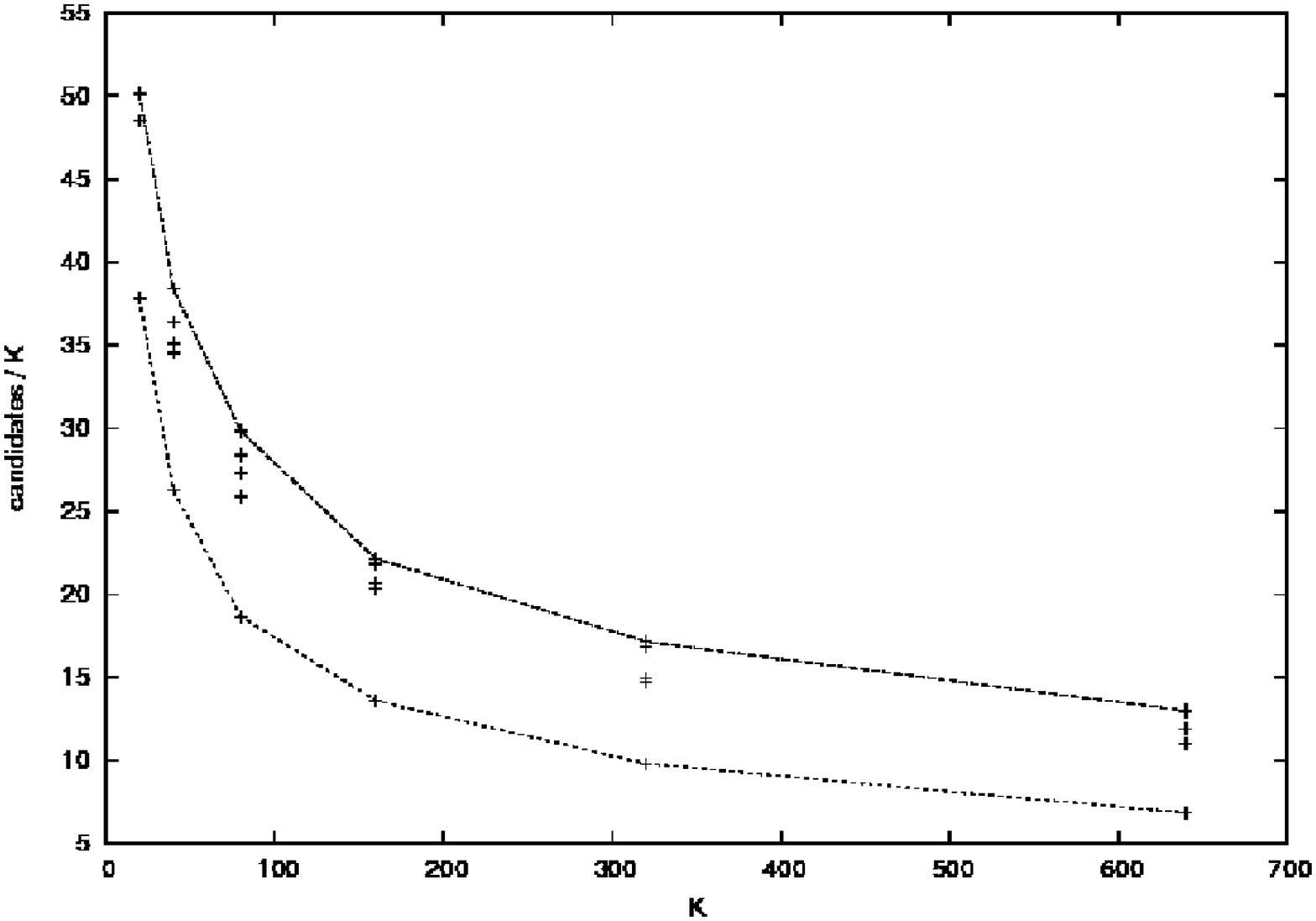}\\
  \caption %
  {Number of candidate nodes visited per neighbor found versus the
  number of K nearest neighbors. Upper and lower curves matches the maximum
  and minimum candidates found.}\label{fig.timing.2}
  } %
\end{figure}

The time complexity of the top down kNN search shown in Algorithm~1 is
measured by counting the total number of nodes visited and the total number of
insertions of unit nodes into the kNN list per query. This latter corresponds
to the number of individual points which were inserted into the kNN list
regardless they remained in the list until the search ended. The number of node
candidates is greater than the desired number $K$ of nearest neighbors of the
query point.

There are two critical costs on the present search complexity. The first
one concentrates on the total number of internal nodes visited along the entire
search per query (Figure~\ref{fig.timing.1}). The other one is the fraction of
visited nodes which has in fact a kNN candidate -- it means that the more
efficient is the search the smaller is the proportional number of rejected nodes
(Figure~\ref{fig.timing.2}).

Examining Figure~\ref{fig.timing.2} one may see that the search efficiency
increases with the increase of the prefixed number $K$ of the
nearest neighbors found. The plausible explanation for this speedup is that the
$K/N$ ratio is roughly the probability of finding
some of the $K$ nearest neighbors of a given query point.

The search complexity
depends on the data distribution, so that in low contrast regions the time
complexity is smaller than finding the kNN in higher contrast distributions.
This effect is responsible for the dispersion between the upper and lower
bounding curves shown in both figures~\ref{fig.timing.1} and \ref{fig.timing.2}.

\section{Application. Anisotropic interpolation}
\label{sec:application}

To illustrate one of a number of purposes of the covariance quadtree
method, we show an image reconstruction from a noise image, as in the
data points shown in Figure~\ref{fig.m51.points.1}. The reconstruction is
performed using a density estimation technique based on compact support
kernels. Usually, a kernel is a good function, which is a member from
a Dirac's $\delta$-sequence.

The adopted kernel is a radial basis function, whose support radius
is set by the distance from the origin to the outermost point from the
K-nearest neighbors. Adopting the Mahalanobis distance, the kernel
support is an ellipsoid whose semi-major axes are defined by the square
root of the principal components of the covariance matrix estimated for
the K-nearest neighbors.

The anisotropic density is estimated from the following summation:
\begin{equation}\label{eq:anisointerp}
\rho(\vec x)=\frac1{N}\sum_{j}
    \frac{K(\xi_j(\vec{x}))}{\sigma_a\sigma_b}
\end{equation}
where the metric function $\xi_j$ is given by
\begin{equation}
\xi_j^2(\vec{x})=
		\frac{|(\vec{x}-\vec{x}_j)^{\rm T}\vec{u}_a|^ 2}{\sigma_a^2}
		+
    \frac{|(\vec{x}-\vec{x}_j)^{\rm T}\vec{u}_b|^ 2}{\sigma_b^2}
\end{equation}
$\vec{u}_a$, $\vec{u}_b$ are the two-dimensional eigenvectors,
${\sigma_a^2}$, ${\sigma_b^2}$ are the eigenvalues, $\vec{x}$
is the query point, in which the estimation is made, and $\vec{x}^\prime$ is
the neighboring position. The kernel function $K$ is normalized in order to
give
\begin{equation}
\int\rho(\vec x)\mathrm{d}x^2=1.
\end{equation}

\begin{figure}
  \center{
  \includegraphics[width=100mm,height=67.5mm]{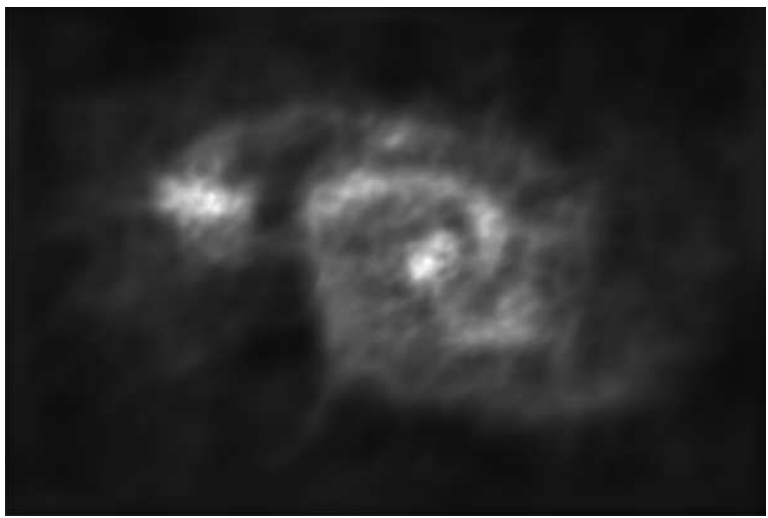}\\
  \caption{The anisotropically interpolated image of the%
  M51 galaxy, using only 4,401 randomly chosen points from the 44,081 shown
  in Figure~\ref{fig.m51.points.1}.}\label{fig.m51noiseAI}
  } %
\end{figure}

To illustrate the anisotropic density estimation we recall the examples
previously given with a random distribution of 44,081 points drawn from the
gray-scale image of the spiral system M51. The integral given in
equation~(\ref{eq:anisointerp}) is easily translated to a grid image so that
densities are normalized to comprise the 8-bit color range of the output
image in Figure~\ref{fig.m51noiseAI}. The density estimation shown in
Figure~\ref{fig.m51noiseAI} was performed over 4,401 ($\sim10\%$) of the points
shown previously in Figure~\ref{fig.m51.points.1}, which reduced considerably
the computational cost of the gray-scale construction.

The integral in (\ref{eq:anisointerp}) is effectively performed over the
kernel support, which is the ellipsoidal region encompassing the $K=200$ nearest
neighbors found with the method described in Subsection~\ref{sec:3.2}. The
adopted kernel model to perform the density estimation shown in
Figure~\ref{fig.m51noiseAI} was the cubic B-spline, largely used in the SPH
literature \cite[e.g., ][]{1977MNRAS.181..375G,Owen+Villumsen+Shapiro+Martel1998,Pelupessy+Schaap+van_de_Weygaert2003}.

\section{Conclusion}
\label{sec:5}

The present work is a preliminary approach to introduce and validate a
fast algorithm to perform anisotropic search for the
exact k nearest neighbors. The adopted concept of anisotropy was based on the
Gaussian multivariate interpretation of the local data distribution assigned to
the covariance quadtree\ nodes. The method detects spatial structures by
conforming the kNN ellipsoid to the principal directions of the local data dispersion.

Differently from the work of \cite{770370}, the modified query
to node distance depends not only on the principal component but on intermediate
covariance component which gives the the maximum contribution to the distance
estimation, which improves the search on choosing the closest covariance cell
(covariance quadtree's node) in a more refined criterion, avoiding deeper
descents along the kNN search, typical of strictly binary trees.

Both the total performance and the search efficiency are approximately
logarithmic per search if the number of neighbors $K$ is relatively large
($K\sim 2\sqrt{N}$). The search efficiency was measured as the ratio of the
number of definitive neighbors to the number of insertions $C$ in the kNN list.
The speedup with the number $K$ of neighbors can be explained by the increase on
the probability $\sim{K/C}$ that an insertion in the kNN list is definitive.

We tested the present method with a simple application of the
gray-scale image retrieval from a noise data distribution, which revealed a
good quality image even using only 10\% of the data points.

The proposed technique might be extended to the case of any other
positive-definite bilinear form. For instance the inertia tensor might be used
to define the principal directions to divide a solid into small pieces separated
from each other by normal cut planes. In this case, if the solid were
represented by particles, as in the SPH scheme, the anisotropic search should
be performed using the inverse inertia tensor in place of the covariance matrix
modifying the Mahalanobis metric in the Algorithm~1.

Future applications of the present anisotropic kNN approach shall be
concentrated in the investigation of SPH simulation of strongly compressive
regimes, in which the adopted metric tensor presently used in the Mahalanobis
distance must be replaced by the stress tensor.


\bibliography{kNNref}

\begin{thebibliography}{24}
\providecommand{\natexlab}[1]{#1}
\expandafter\ifx\csname urlstyle\endcsname\relax
  \providecommand{\doi}[1]{doi:\discretionary{}{}{}#1}\else
  \providecommand{\doi}{doi:\discretionary{}{}{}\begingroup
  \urlstyle{rm}\Url}\fi

\bibitem[{Bishop(1995)}]{Bishop1995}
Bishop C.M.
\newblock \emph{{Neural Networks for Pattern Recognition}}.
\newblock {Oxford} {University} {Press}, Bookcraft {(Bath)} {Ltd.}, {Midsomer
  Norton}, {Avon}, {Great Britain}, 1995.

\bibitem[{Burschka et~al.(2004)Burschka, Li, Taylor, \amcabtxand{}
  Hager}]{IROS04-BuHa}
Burschka D., Li M., Taylor R., \amcabtxand{} Hager G.D.
\newblock {S}cale-{I}nvariant {R}egistration of {M}onocular {S}tereo {I}mages
  to 3{D} {S}urface {M}odels.
\newblock In \emph{Proceedings of 2004 IEEE/RSJ International Conference on
  Intelligent Robots and Systems, September 28 - October 2, 2004, Sendai,
  Japan}, \amcabtxpages{} 2581--2586. IEEE, 2004.

\bibitem[{D'haes et~al.(2003)D'haes, van Dyck, \amcabtxand{} Rodet}]{770370}
D'haes W., van Dyck D., \amcabtxand{} Rodet X.
\newblock Pca-based branch and bound search algorithms for computing k nearest
  neighbors.
\newblock \emph{Pattern Recogn. Lett.}, 24(9-10):1437--1451, 2003.
\newblock ISSN 0167-8655.
\newblock \doi{http://dx.doi.org/10.1016/S0167-8655(02)00384-7}.

\bibitem[{Duda \amcabtxand{} Hart(1973)}]{Duda}
Duda R.O. \amcabtxand{} Hart P.E.
\newblock \emph{{Pattern Classification and Scene Analysis}}.
\newblock {New York: Wiley \& Sons}, 1973.

\bibitem[{Finkel \amcabtxand{} Bentley(1974)}]{FinkelBentley1974}
Finkel R. \amcabtxand{} Bentley J.
\newblock Quad trees: A data structure for retrieval on composite keys.
\newblock \emph{Acta Informatica}, 4 (1)(1):1–9, 1974.
\newblock \doi{10.1007/BF00288933}.

\bibitem[{Gingold \amcabtxand{} Monaghan(1977)}]{1977MNRAS.181..375G}
Gingold R.A. \amcabtxand{} Monaghan J.J.
\newblock {Smoothed particle hydrodynamics - Theory and application to
  non-spherical stars}.
\newblock \emph{{M}onthly {N}oticies of the {R}oyal {A}stronomical {S}ociety},
  181:375--389, 1977.

\bibitem[{Jiang et~al.(2006)Jiang, Oliveira, \amcabtxand{}
  Sousa}]{Jiang+Oliveira+Sousa2006}
Jiang F., Oliveira M.S., \amcabtxand{} Sousa A.C.
\newblock Sph simulation of transition to turbulence for planar shear flow
  subjected to a streamwise magnetic field.
\newblock \emph{Journal of Computational Physics}, 217:485--501, 2006.
\newblock \doi{10.1016/j.jcp.2006.01.009}.

\bibitem[{Jollife(1986)}]{Jollife1986}
Jollife I.T.
\newblock \emph{{Principal Component Analysis}}.
\newblock {New York:} {Springer-Verlag}, 1986.

\bibitem[{Liu et~al.(2006)Liu, Liu, \amcabtxand{} Lam}]{Liu+Liu+Lam2006}
Liu M.B., Liu G.R., \amcabtxand{} Lam K.Y.
\newblock Adaptive smoothed particle hydrodynamics for high strain
  hydrodynamics with material strength.
\newblock \emph{Shock Waves}, 15(1):21--29, 2006.

\bibitem[{Lucy(1977)}]{1977AJ.....82.1013L}
Lucy L.B.
\newblock {A numerical approach to the testing of the fission hypothesis}.
\newblock \emph{{A}stronomical {J}ournal}, 82:1013--1024, 1977.

\bibitem[{Ma \amcabtxand{} Ji(1998)}]{ma98modeling}
Ma S. \amcabtxand{} Ji C.
\newblock Modeling video traffic in the wavelet domain.
\newblock In \emph{{INFOCOM} (1)}, \amcabtxpages{} 201--208. IEEE, 1998.

\bibitem[{Mahalanobis(1936)}]{Mahalanobis1936}
Mahalanobis P.
\newblock On the generalized distance in statistics.
\newblock \emph{National Institute of Science in India}, 12:49--55, 1936.

\bibitem[{Marinho et~al.(2001)Marinho, Andreazza, \amcabtxand{}
  L{\'{e}}pine}]{Marinho+Andreazza+Lepine2001}
Marinho E.P., Andreazza C.M., \amcabtxand{} L{\'{e}}pine J.R.D.
\newblock {SPH simulations of clumps formation by dissipative collisions of
  molecular clouds. II. Magnetic case}.
\newblock \emph{{A}stronomy and {A}strophysics}, 379:1123--1137, 2001.
\newblock \doi{10.1051/0004-6361:20011352}.

\bibitem[{Marinho \amcabtxand{} L{\'{e}}pine(2000)}]{Marinho+Lepine2000}
Marinho E.P. \amcabtxand{} L{\'{e}}pine J.R.D.
\newblock {SPH simulations of clumps formation by dissipative collision of
  molecular clouds. I. Non magnetic case}.
\newblock \emph{{A}stronomy and {A}strophysics {S}upplement}, 142:165--179,
  2000.

\bibitem[{McNames(2001)}]{mcnames01fast}
McNames J.
\newblock A fast nearest-neighbor algorithm based on a principal axis search
  tree.
\newblock \emph{IEEE Transactions on Pattern Analysis and Machine
  Intelligence}, 23(9):964--976, 2001.

\bibitem[{Minasny et~al.(2007)Minasny, McBratney, \amcabtxand{}
  Walvoort}]{Minasny2007}
Minasny B., McBratney A.B., \amcabtxand{} Walvoort D.
\newblock The variance quadtree algorithm: Use for spatial sampling design.
\newblock \emph{Computers \& Geosciences}, 33:383–392, 2007.

\bibitem[{Mouratidis et~al.(2005)Mouratidis, Papadias, \amcabtxand{}
  Tao}]{1092803}
Mouratidis K., Papadias D., \amcabtxand{} Tao Y.
\newblock A threshold-based algorithm for continuous monitoring of k nearest
  neighbors.
\newblock \emph{IEEE Transactions on Knowledge and Data Engineering},
  17(11):1451--1464, 2005.
\newblock ISSN 1041-4347.
\newblock \doi{http://dx.doi.org/10.1109/TKDE.2005.172}.
\newblock Member-Spiridon Bakiras.

\bibitem[{Owen et~al.(1998)Owen, Villumsen, Shapiro, \amcabtxand{}
  Martel}]{Owen+Villumsen+Shapiro+Martel1998}
Owen J.M., Villumsen J.V., Shapiro P.R., \amcabtxand{} Martel H.
\newblock {Adaptive Smoothed Particle Hydrodynamics: Methodology. II.}
\newblock \emph{{A}strophysical {J}ournal {S}upplement}, 116:155--+, 1998.
\newblock \doi{10.1086/313100}.

\bibitem[{Pelupessy et~al.(2003)Pelupessy, Schaap, \amcabtxand{} van~de
  Weygaert}]{Pelupessy+Schaap+van_de_Weygaert2003}
Pelupessy F.I., Schaap W.E., \amcabtxand{} van~de Weygaert R.
\newblock Density estimators in particle hydrodynamics. dtfe versus regular
  sph.
\newblock \emph{Astronomy \& Astrophysics}, 403:389--398, 2003.
\newblock \doi{10.1051/0004-6361:20030314}.

\bibitem[{Rehrmann \amcabtxand{} Priese(1998)}]{DBLP:conf/accv/RehrmannP98}
Rehrmann V. \amcabtxand{} Priese L.
\newblock Fast and robust segmentation of natural color scenes.
\newblock In \emph{ACCV (1)}, \amcabtxpages{} 598--606. 1998.

\bibitem[{Ribeiro et~al.(2006)Ribeiro, Riedi, \amcabtxand{} Baraniuk}]{Rib2006}
Ribeiro V.J., Riedi R.H., \amcabtxand{} Baraniuk R.G.
\newblock Optimal sampling strategies for multiscale stochastic processes.
\newblock In \emph{Institute of Mathematical Statistics Lecture Notes -
  Monograph Series}, \amcabtxpages{} 1--29. IEEE, 2006.

\bibitem[{Tran et~al.(2006)Tran, Wehrens, \amcabtxand{}
  Buydens}]{2006TWB.51.513-525}
Tran T.N., Wehrens R., \amcabtxand{} Buydens L.M.
\newblock Knn-kernel density-based clustering for high-dimensional multivariate
  data.
\newblock \emph{Computational Statisticas \& Data Analysis}, 51:513--525, 2006.

\bibitem[{Xu et~al.(2009)Xu, Stansby, \amcabtxand{} Laurence}]{Xu2009}
Xu R., Stansby P., \amcabtxand{} Laurence D.
\newblock Accuracy and stability in incompressible sph (isph) based on the
  projection method and a new approach.
\newblock \emph{Journal of Computational Physics}, In Press, Accepted
  Manuscript:--, 2009.
\newblock ISSN 0021-9991.
\newblock \doi{DOI: 10.1016/j.jcp.2009.05.032}.

\bibitem[{Yu et~al.(2001)Yu, Sharma, Meng, \amcabtxand{} Qin}]{379504}
Yu C., Sharma P., Meng W., \amcabtxand{} Qin Y.
\newblock Database selection for processing k nearest neighbors queries in
  distributed environments.
\newblock In \emph{JCDL '01: Proceedings of the 1st ACM/IEEE-CS joint
  conference on Digital libraries}, \amcabtxpages{} 215--222. ACM Press, New
  York, NY, USA, 2001.
\newblock ISBN 1-58113-345-6.
\newblock \doi{http://doi.acm.org/10.1145/379437.379504}.

\end{thebibliography}
\end{document}